\documentclass[9pt,technote]{IEEEtran}

\usepackage{graphicx}
\usepackage{amsmath}
\usepackage{amssymb}

\usepackage{cs}
\usepackage{mathsymb}

\def\eg{{\em e.g.}}

\def\psd{{p.s.d.}\xspace}

\def\triangle{\delta}

\def\ell{\lambda}

\def\T{{\!\top}}

\graphicspath{{./}{./Fig/}{./Figs/}}

\usepackage{fixltx2e}
\usepackage{cite}

\usepackage{amsmath}
\usepackage{algorithm2e}
\let\savedalgorithm\algorithm
\let\savedendalgorithm\endalgorithm
\newenvironment{algorithmic}{%
\savedalgorithm
}{
\savedendalgorithm
}

\usepackage{algorithm}

\begin{document}

\title{Scalable Large-Margin Mahalanobis Distance Metric Learning}

\author{Chunhua Shen, Junae Kim, and Lei Wang
\thanks
{ 
Manuscript received April X, 200X; revised March X, 200X.
NICTA is funded by the Australian Government as represented by the Department of Broadband,
Communications and the Digital Economy and 
the Australian Research Council through the ICT Center of
Excellence program.
}
\thanks{
C. Shen is with NICTA, Canberra Research Laboratory, 
Locked Bag 8001, Canberra, ACT 2601, Australia,
and also with the Australian National University, Canberra,
ACT 0200,Australia
(e-mail: chunhua.shen@nicta.com.au).
}
\thanks{
J. Kim and L. Wang are with the 
Australian National University, Canberra, ACT 0200, Australia
(e-mail: \{junae.kim, lei.wang\}@anu.edu.au). 
}
\thanks{
Color versions of one or more of the figures in this brief are available online
at http://ieeexplore.ieee.org.
}
\thanks{
Digital Object Identifier 10.1109/TNN.200X.XXXXXXX
}
}

\markboth{IEEE TRANSACTIONS ON NEURAL NETWORKS, VOL.~30, NO.~9, SEPTEMBER 200X}
{SHEN \MakeLowercase{\it et al.}: Scalable Large-Margin Mahalanobis Distance Metric Learning}

\maketitle

\begin{abstract}
    
%
%
%
%
%

    For many machine learning algorithms such as $k$-Nearest Neighbor ($k$-NN) classifiers and $ k
    $-means clustering, often their success heavily depends on the metric used to calculate
    distances between different data points.
    An effective solution for defining such a metric is to
    learn it from a set of labeled training samples.  In this work, we propose a fast and scalable
    algorithm to learn a Mahalanobis distance metric.  The Mahalanobis metric can be viewed as the
    Euclidean distance metric on the input data that have been linearly transformed. 
    By employing
    the principle of margin maximization to achieve better generalization performances, this
    algorithm formulates the metric learning as a convex optimization problem and a positive
    semidefinite (\psd) matrix is the unknown variable. 
    Based on an important theorem that a \psd trace-one matrix
    can always be represented as a convex combination of multiple rank-one matrices, our algorithm
    accommodates any differentiable loss function and solves the resulting optimization problem
    using a specialized gradient descent procedure.  During the course of optimization, the proposed
    algorithm maintains the positive semidefiniteness of the matrix
    variable that is essential for a Mahalanobis metric.
    %
    %
    Compared with conventional methods like standard interior-point algorithms 
    \cite{Boyd2004} or
    the special solver used in Large Margin Nearest Neighbor (LMNN) \cite{Kilian},
    our algorithm is much more efficient and has a better performance in scalability.
    Experiments on benchmark data sets suggest that, compared with
    state-of-the-art
    metric learning algorithms, our algorithm can achieve 
    a comparable classification accuracy with  
    reduced computational complexity.

\end{abstract}


\section{Introduction}

        In many machine learning problems, the distance metric used over the input data has critical
        impact on the success of a learning algorithm.  For instance, $ k $-Nearest Neighbor ($ k
        $-NN) classification \cite{knn}, and clustering algorithms such as $k$-means rely on if an
        appropriate distance metric is used to faithfully model the underlying relationships
        between the input data points.  
        A more concrete example is visual object recognition.
        Many visual recognition tasks can be viewed as inferring a distance metric that
        is able to measure the (dis)similarity of the input visual data, 
        ideally being consistent with human
        perception. Typical examples include  object categorization 
        \cite{Winn-ICCV-05} and
        content-based image retrieval \cite{Smeulders-PAMI-2000}, in which
        a similarity metric is needed to
        discriminate different object classes or 
        relevant and irrelevant images against a given query.
        %
        %
      As one of the most classic and simplest classifiers, $k$-NN has
      been applied to a wide range of vision tasks and it is the
      classifier that directly depends on a predefined distance metric. 
      An appropriate distance
      metric is usually needed for achieving a promising accuracy.
      Previous work (\eg, \cite{Xing,Yang08})
      has shown that compared to
      using the standard Euclidean distance, applying an well-designed distance
      often can significantly boost the classification accuracy of a $k$-NN 
      classifier.
      In this work, we propose a scalable and fast algorithm to 
      learn a Mahalanobis distance metric.
      Mahalanobis metric removes the main limitation of the Euclidean metric
      in that
      it corrects for correlation between the different features.
      
%
%
      Recently, much research effort has been spent on learning
      a Mahalanobis distance metric from labeled data \cite{Xing,Yang08,Kilian,
      Davis07}. Typically, a convex cost function is defined such that 
      a global optimum can be achieved in polynomial time.
      It has been shown in the statistical learning theory 
      \cite{Vapniklearnig} that
      increasing the margin between different classes helps to reduce the
      generalization error. Inspired by the work of \cite{Kilian},
      we directly learn the Mahalanobis matrix from a set of 
      {\em distance comparisons},
      and optimize it via margin
      maximization. The intuition is that 
      such a learned Mahalanobis distance metric may achieve sufficient
      separation at the boundaries
      between different classes. More importantly, 
      we address the scalability problem of learning the Mahalanobis
      distance matrix in the presence of high-dimensional feature
      vectors, which is a critical issue of distance metric learning. 
      As indicated in a theorem in 
      \cite{Shen2008psd},
      a positive semidefinite trace-one matrix can always 
      be decomposed as a
      convex combination of a set of rank-one matrices. This
      theorem has inspired us to develop a fast optimization
      algorithm that works in the style of gradient descent. At each iteration,
      it only needs to find the  principal eigenvector of a matrix of size
      ${D \times D} $ ($ D$ is the dimensionality of the input data)
      and a simple matrix update. This process
      incurs much less computational overhead than the metric learning 
      algorithms in the literature
      \cite{Kilian,Boyd2004}.
      Moreover, thanks to the above
      theorem, this process automatically preserves the \psd
      property of the Mahalanobis matrix. To verify
      its effectiveness and efficiency, the proposed algorithm is tested on a few benchmark data
      sets and is compared with the state-of-the-art distance metric learning algorithms.
      As experimentally demonstrated, $k$-NN with the
      Mahalanobis distance learned by our algorithms attains comparable (sometimes slightly 
      better) 
      classification accuracy. Meanwhile, in terms of the computation time,
      the proposed algorithm
      has much better scalability in terms of the 
      dimensionality of input feature vectors.

    We briefly review some related work before we present our work.
    Given a classification task, some previous 
    work on learning a distance metric aims to find a
    metric that makes the data in the same class close and
    separates those in different classes from each other as far as possible.
    Xing \etal \cite{Xing} proposed an approach to learn a Mahalanobis
    distance for supervised clustering. It minimizes the sum of the distances
    among data in the same class while maximizing
    the sum of the distances among data in different classes. Their work shows that
    the learned metric could improve clustering performance significantly. However,
    to maintain the \psd property, they have used projected gradient descent and 
    their approach has to
    perform a {\em full} eigen-decomposition of the Mahalanobis matrix at each
    iteration. Its computational cost rises rapidly when the number of
    features increases, and this makes it less efficient in coping with
    high-dimensional data. 
    Goldberger \etal \cite{Jacob} developed an algorithm termed
    Neighborhood Component Analysis (NCA), which learns a
    Mahalanobis distance by minimizing the leave-one-out cross-validation
    error of the $k$-NN classifier on the training set.
    NCA needs to solve a
    non-convex optimization problem, which might have many local optima.
    Thus it is critically important to start the search from a reasonable 
    initial point. Goldberger \etal have used the result of linear discriminant
    analysis as the initial point.
    In NCA, the variable to optimize is the projection matrix.

    The work closest to ours is Large Margin Nearest Neighbor (LMNN)
    \cite{Kilian} in the sense that it
    also learns a Mahalanobis distance in the large margin framework.
    In their approach, the distances between each sample and its 
    ``target neighbors'' are minimized
    while the distances among the data with different labels are maximized.
    A convex objective function is obtained and the resulting problem is
    a semidefinite program (SDP). Since conventional interior-point  based
    SDP solvers can only solve problems of up to a few thousand variables, 
    LMNN has
    adopted an alternating projection algorithm for solving the 
    SDP problem. At each iteration, 
    similar to \cite{Xing}, also a full eigen-decomposition is needed. 
    Our approach is largely inspired by their work.
    %
    %
    Our work differs LMNN \cite{Kilian} in the following: 
    (1) LMNN learns the metric from the pairwise distance information.  
       In contrast, our algorithm uses examples of proximity comparisons among
       triples of objects (\eg, example $i$ is closer to example $j$ than example $k$).
       In some applications like image retrieval, this type of information could be easier to 
       obtain than to tag the actual class label of each training image.
       Rosales and Fung \cite{Rosales2006} have used similar ideas on metric learning;
    (2)    
        More importantly, we design an optimization method that
        has a clear advantage on 
        computational efficiency (we only
        need to compute the leading eigenvector at each iteration).
        The optimization problems of \cite{Kilian} and \cite{Rosales2006}
        are both SDPs, which are computationally heavy. 
        Linear programs (LPs) are used in \cite{Rosales2006} 
        to approximate the SDP problem. 
        It remains unclear how well this approximation is.

        The problem of learning a kernel from a set of labeled data shares similarities with 
        metric learning because the optimization involved
        has similar formulations. 
        Lanckriet \etal \cite{lanckriet} and Kulis \etal \cite{Kulis2009}
        considered learning \psd kernels subject to some pre-defined constraints.  
        An appropriate kernel can often offer algorithmic improvements. It
        is possible to apply the proposed gradient descent optimization technique
        to solve the kernel learning problems. We leave this topic for future study.

        The rest of the paper is organized as follows.
        Section~\ref{Sec:kNNMM}
        presents the convex formulation of learning a Mahalanobis 
        metric. In
        Section~\ref{Sec:kNNMM-SDP}, we show how to efficiently solve
        the optimization problem by a specialized 
        gradient descent procedure, which is the main contribution of
        this work.
        The performance of our approach is
        experimentally demonstrated in Section~\ref{Sec:experiments}.
        Finally, we conclude 
        this work in Section~\ref{Sec:conclusion}.

\section{Large-Margin Mahalanobis Metric Learning}

\label{Sec:kNNMM}

    In this section, we propose our distance metric learning approach as follows.
    The intuition is to find a particular distance metric for which the
    margin of separation between the classes is maximized.
    In particular, we are interested in learning a quadratic Mahalanobis metric.

    Let $\mathbf a_i \in \mathbb R^D (i=1,2,\cdots, n)$ denote a training
    sample where $ n $ is the number of training samples and $D$ is the
    number of features.
    To learn a Mahalanobis distance, we create a set $\mathcal S$ that
    contains a group of
    training triplets as $\mathcal S = \{(\mathbf a_i, \mathbf a_j, \mathbf
    a_k)\}$, where $\mathbf a_i$
    and $\mathbf a_j$ come from the same class and $\mathbf a_k$
    belongs to different classes. A Mahalanobis distance is defined as follows. 
    Let $\mathbf P \in \mathbb R^{D \times d}$ denote a linear
    transformation and $\mathbf{dist}$ be the squared Euclidean distance in the transformed
    space. The squared distance between the projections of $\mathbf a_i$
    and $\mathbf a_j$ writes:
    \begin{equation}
    \mathbf{dist}_{ij}=\|\mathbf P^{\T} \mathbf a_i - \mathbf P^{\T} \mathbf
    a_j\|^2_2 = (\mathbf a_i - \mathbf a_j)^{\T} \mathbf P \mathbf P^{\T} (\mathbf a_i - \mathbf
    a_j).
    \end{equation} 
    According to the class memberships of $\mathbf a_i$, $\mathbf a_j$ and $\mathbf a_k$,
    we wish to achieve $\mathbf{dist}_{ik}\geq\mathbf{dist}_{ij}$ and it can be obtained as
    \begin{equation}
    (\mathbf a_i - \mathbf a_k)^{\T} 
    \mathbf P \mathbf P^{\T} 
    (\mathbf a_i - \mathbf a_k)\geq (\mathbf a_i - \mathbf a_j)^{\T} \mathbf P \mathbf P^{\T} 
    (\mathbf a_i - \mathbf a_j).
    \label{EQ:3}
    \end{equation}
    It is not difficult to see that this inequality is generally not a convex constrain in $\mathbf P$
    because the difference of quadratic terms in $\mathbf P$ is involved. In order to make this
    inequality constrain convex, a new variable $\mathbf X = \mathbf P
    \mathbf P^{\T}$ is introduced and used throughout the whole learning process.
    Learning a Mahalanobis distance is essentially learning the Mahalanobis matrix $\mathbf X$.
    \eqref{EQ:3} becomes linear in $ \bf X $. 
    This is a typical technique to {\em convexify} a problem in convex optimization
    \cite{Boyd2004}.

    \subsection{Maximization of a soft margin}
    \label{Sec:soft_margin}

    In our algorithm, a \textit{margin} is
    defined as the difference between $\mathbf{dist}_{ik}$ and
    $\mathbf{dist}_{ij}$, that is,
    \begin{equation}
    \begin{array}{ll}
        \rho_r = (\mathbf a_i - \mathbf a_k)^{\T} {\mathbf X}
        (
    \mathbf a_i - \mathbf a_k) - (\mathbf a_i - \mathbf a_j)^{\T}
    {\mathbf X} ( \mathbf a_i - \mathbf a_j),\\
     \forall (\mathbf a_i, \mathbf a_j, \mathbf a_k) \in \mathcal
    S,~~r=1,2,\cdots,|\mathcal S|. 
    \end{array}
    \label{EQ:A1}
    \end{equation}
    Similar to the large margin principle that has been widely used
    in machine learning algorithms such as
    support vector machines and boosting,  
    here we maximize this margin \eqref{EQ:A1} to obtain 
    the optimal Mahalanobis matrix $\mathbf X$. 
    Clearly, the larger is the margin $ \rho_r $, the better metric might be
    achieved. 
    %
    %
    To enable some flexibility, \ie, 
    to allow some inequalities of \eqref{EQ:3} not 
    to be satisfied, 
    a soft-margin criterion is needed. 
    Considering these factors, we could define
    the objective function for learning $\mathbf X$ as
    \begin{equation}
        \label{eqn:SDP-optimization}
    \begin{array}{lll}
    &\!\! \max_{\rho, \mathbf X, \boldsymbol\xi}
                           \quad
                           {\rho - C\sum_{r=1}^{|\mathcal S|}\xi_r},
    \,\, \st \\
    &  \mathbf X \succcurlyeq 0, \mathbf {Tr}(\mathbf X) = 1, \\
    &                    
    \xi_r \geq 0,~~r=1,2,\cdots,|\mathcal S|,\\
    & (\mathbf a_i - \mathbf a_k)^{\T} \mathbf X ( \mathbf a_i - \mathbf a_k) -
    (\mathbf a_i - \mathbf a_j)^{\T} \mathbf X ( \mathbf a_i - \mathbf a_j) \geq \rho - \xi_r,
    \\           
    &   \forall (\mathbf a_i, \mathbf a_j, \mathbf a_k) \in \mathcal S,
    \end{array}
    \end{equation}
    where $\mathbf X \succcurlyeq 0$ constrains $\mathbf X$ to be a \psd matrix and
    $\mathbf {Tr}(\mathbf X)$ denotes the trace of $\mathbf X$.
    $r$ indexes the training set $\mathcal S$ and $|\mathcal S|$ denotes the size of $\mathcal S$.
    $C$ is an algorithmic parameter that balances the violation of \eqref{EQ:3}
    and the margin maximization. $\xi \geq 0$ is the slack
    variable similar to that used in support vector machines
    and it corresponds to the soft-margin hinge loss.
    Enforcing $\mathbf {Tr}(\mathbf X) = 1$ removes
    the scale ambiguity because the inequality constrains are scale
    invariant. 
    To simplify exposition, we define
    \begin{equation}
    \mathbf A^r = (\mathbf a_i - \mathbf a_k)( \mathbf a_i - \mathbf a_k)^{\T} -
    (\mathbf a_i - \mathbf a_j)( \mathbf a_i - \mathbf
    a_j)^{\T}.
    \end{equation}ssss
    Therefore, the last constraint in \eqref{eqn:SDP-optimization} can be written as
    \begin{equation}
    \big<\mathbf A^r,\mathbf X\big> \geq \rho - \xi_r, \quad r=1,\cdots,|S|. 
    \end{equation}
    Note that this is a linear constrain on $\mathbf X$. Problem
    \eqref{eqn:SDP-optimization} is thus a typical SDP problem since it has a
    linear objective function and linear constraints plus a \psd
    conic constraint. One may solve it using off-the-shelf SDP solvers
    like CSDP \cite{Borchers1999CSDP}.
    However, directly solving the problem
    \eqref{eqn:SDP-optimization} using those standard interior-point SDP solvers
    would quickly become computationally intractable with the increasing
    dimensionality of feature vectors. 
%
    We show how to efficiently solve \eqref{eqn:SDP-optimization} in a 
    fashion of first-order gradient descent. 


\subsection{Employment of a differentiable loss function}
\label{Sec:loss-function}

    It is proved in~\cite{Shen2008psd} that \textit{a \psd matrix can always be decomposed as a
    linear
    convex
    combination of a set of rank-one matrices}. 
    In the context of our problem, this means that
    $\mathbf X=\sum_i\theta_{i}\mathbf Z_{i}$, where $\mathbf Z_i$ is a rank-one matrix and $\mathbf
    {Tr}(\mathbf Z_i) = 1$. This important result inspires us to develop a gradient descent based
    optimization algorithm. In each iteration, $\mathbf X$ can be updated as
    \begin{equation}
        \mathbf X_{i+1}=\mathbf X_{i}+\alpha(\triangle {\mathbf X} - {\mathbf X}_i) = \mathbf X_{i}+\alpha
    \mathbf p_i,~~~0\leq\alpha\leq 1,
    \end{equation}
    where $\triangle\mathbf X$ is a rank-one and trace-one matrix. 
    $\mathbf p_i$ is the search direction. 
    It is straightforward to verify that
    $ { \bf Tr } ( { \bf X}_{i+1} ) = 1 $, and $ {\bf X}_{i+1} \succcurlyeq 0$ hold.
    This is the starting point of our gradient descent algorithm. 
    With this update strategy, the trace-one and positive semidefinteness of $ \bf X $
    is always retained. 
    We show how to calculate this search direction
    in
    Algorithm~\ref{ALG:2}.
    Although it is possible to use subgradient methods to optimize non-smooth objective functions,
    we use a differentiable objective function instead so that the optimization procedure
    is simplified (standard gradient descent can be applied). 
    %
    %
    So,  we need to ensure that the objective  function is
    differentiable with respect to the variables $\rho$ and $\mathbf X$.

         Let $f(\cdot)$ denote the objective function and $\ell (\cdot)$ be a
         loss function. Our objective function can be rewritten as
    \begin{equation}
        \label{eqn:f}
    f(\mathbf X, \rho) = \rho - C \cdot \sum_{r=1}^{|\mathcal S|}{ \ell \left(
                 \big<\mathbf A^r,\mathbf
    X\big> - \rho \right)}.
    \end{equation}
    The above problem \eqref{eqn:SDP-optimization} adopts the hinge loss function that is defined
    as $ \ell(z) = \max(0, -z)$. However, the hinge loss is not differentiable
    at the point of $z = 0$, and
    standard gradient-based optimization cam be applied directly. 
    In order to make standard gradient descent methods applicable, we propose
    to use differentiable loss functions, 
    for example, the squared hinge loss or Huber loss
    functions as discussed below.

        \begin{figure}[t!]
        \centering
        \includegraphics[width=0.45\textwidth]{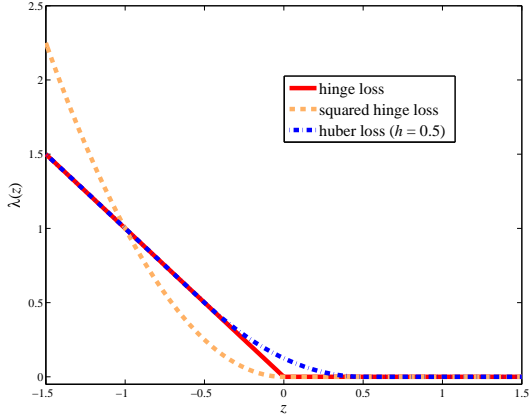}\\
        \caption{The hinge loss, squared hinge loss and Huber loss.}
        \label{fig:loss}
        \end{figure}


    The squared hinge loss function can be represented as
    \begin{align}
        \label{eqn:squaredloss}
    & \ell \left(\big<\mathbf A^r,\mathbf X\big> - \rho \right) = 
           \notag
           \\
    & \quad  
    \left\{\begin{array}{ll}
    0, & \mbox{if }  \left(\big<\mathbf A^r,\mathbf X\big> - \rho \right)\ge 0, \\
    \left(\big<\mathbf A^r,\mathbf X\big> - \rho\right)^2, & \mbox{if } \left(\big<\mathbf A^r,
    {\mathbf X}\big> - \rho \right) < 0.
    \end{array}\right.
    \end{align}
    As shown in Fig.~\ref{fig:loss}, this function connects the positive and zero segments
    smoothly and it is differentiable everywhere including the point $z = 0$.
    %
    We also consider the Huber loss function in this work:
    \begin{align}
        \label{eqn:Huberloss}
    & \ell 
    \left( \big<\mathbf A^r,\mathbf X\big> - \rho \right) =   \notag
    \\
    &  \quad  \left\{\begin{array}{ll}
        0, & \mbox{if }\left( 
                       \big<\mathbf A^r,\mathbf X\big> - \rho \right) \ge h, \\
    \frac{ \left( h - \left(  \left< \mathbf A^r,\mathbf X \right> - \rho
                      \right)
            \right)^2}{4h}, & \mbox{if }{-h} < \left(\big<\mathbf A^r,\mathbf
    X\big> - \rho \right) < h, \\\\
    -( \left<\mathbf A^r,\mathbf X \right> - \rho), & \mbox{if } \left(\big<\mathbf A^r,\mathbf
    X\big> - \rho \right) \le -h,
    \end{array}\right.
    \end{align}
    where $h$ is a parameter whose value is usually between $0.01$
    and $0.5$. A Huber loss function with $h=0.5$ is plotted in Fig.~\ref{fig:loss}.
    There are three different parts in the Huber loss function, and they
    together form a continuous and differentiable function. This loss
    function approaches the hinge loss curve when $h \to 0$.
    Although the Huber loss is more complicated than the squared hinge loss, its
    function value increases linearly with the value of $\big<\mathbf A^r,\mathbf X\big> - \rho$.
    Hence, when a training set contains outliers or samples heavily contaminated by noise,
    the Huber loss might give a more reasonable (milder) penalty than the squared hinge loss does.
    We discuss both loss functions in our experimental study. Again, we highlight that by using these two loss
    functions, the cost function $f(\mathbf X,
    \rho)$ that we are going to optimization becomes
    differentiable with respect to both $\mathbf X$ and $\rho$.

%
%



\def\Dot{ { \cdot \,} }

\SetVline \linesnumbered

\begin{algorithm}[t]
\caption{The proposed optimization algorithm.}
\begin{algorithmic}
{
   \KwIn{
   \begin{itemize}
       \item
             The maximum number of iterations $ K $;
       \item
           A pre-set tolerance value $\varepsilon$ (\eg, $ 10^{-5})$.
   \end{itemize}
   }
   { {\bf Initialize}:
        $\mathbf X_0 $ such that $  {\bf Tr}({\bf X}_0) = 1, {\bf rank}( {\bf X}_0) =1 $\;
    }%
\For{ $ k = 1,2,\cdots, K $ } {
$\Dot$
        Compute $\rho_k$ by  solving the subproblem
        $\rho_k = \underset{\rho > 0}{\operatorname{arg\,max}} \, f(\mathbf X_{k-1},
        \rho)$\;

$\Dot$ Compute $\mathbf X_k$ by solving the problem
        $\mathbf X_k = \underset{\mathbf X \succcurlyeq 0,
                    \mathbf {Tr}(\mathbf X) = 1}{\operatorname{arg\,max}} \, f(\mathbf X,
        \rho_k)$\;
$\Dot$
\If{ $ k>1 $ {\rm and} 
        $ |f(\mathbf X_k,\rho_k) - f(\mathbf
        X_{k-1},\rho_{k})| < \varepsilon $ {\rm and} $ |f(\mathbf X_{k-1},\rho_{k}) - f(\mathbf
        X_{k-1},\rho_{k-1})| < \varepsilon $}
        {break (converged)\;}

}
\KwOut{
        The final \psd matrix $ \mathbf X_k $.}
}
\end{algorithmic}
\label{ALG:0}
\end{algorithm}

\SetVline
\linesnumbered

\begin{algorithm}[t]
\caption{Compute $\mathbf{X}_k$
        in the proposed algorithm.}
\begin{algorithmic}
{
   \KwIn{
   \begin{itemize}
       \item
        $\rho_k$ and $ \mathbf{X}_1 $ which is an initial approximation of $ \mathbf{X}_k $;
        \item
        The maximum number of iterations $ J $.
   \end{itemize}
   }
\For{ $ i = 1,2,\cdots, J $ } {
$\Dot$
Compute ${\bf v}_i$ that corresponds to the largest eigenvalue $l_i$ of
        the matrix
        $\nabla f(\mathbf{X}_i, \rho_k)$;\

$\Dot$
        \If{
        $ l_i < \varepsilon $}
        {break (converged)\;}
$\Dot$
        Let the search direction be
        $\mathbf{p}_i = \mathbf v_i \mathbf v_i^{\T} - \mathbf X_i$\;
$\Dot$
        Set $\mathbf X_{i+1} = \mathbf X_{i} +
        \alpha\mathbf{p}_i$. Here $  \alpha $ is found by line search\;
} \KwOut{ Set $\mathbf{X}_{k} = \mathbf X_{i}$. } }
\end{algorithmic}
\label{ALG:2}
\end{algorithm}

   \section{A scalable and fast optimization algorithm}
   \label{Sec:kNNMM-SDP}

    The proposed algorithm maximizes the objective function
    iteratively, and in each iteration the two variables $\mathbf X$ and
    $\rho$ are optimized alternatively.
    Note that the optimization in this alternative strategy retains
    the global optimum  because $f(\mathbf X,\rho)$ is a convex function in both
    variables 
    $(\mathbf X, \rho)$ and $ (\mathbf X, \rho) $ are not coupled together. 
    We summarize the proposed
    algorithm  in Algorithm~\ref{ALG:0}. Note that $\rho_k$ is a scalar and Line 3
    in
    Algorithm \ref{ALG:0}  can be solved directly by a simple
    one-dimensional maximization process. However, $\mathbf X$ is a \psd
    matrix with size of $D \times D$. Recall that $D$ is the dimensionality of
    feature vectors. The following section presents
    how $\mathbf X$ is efficiently optimized in our algorithm.

%
\subsection{Optimizing for the Mahalanobis matrix ${\mathbf X}_k$}
\label{Sec:kNNMM-SDP1}

    Let
    $\mathcal P =\{\mathbf{X}\in \mathbb R^{D \times D}: 
    \mathbf{X}\succcurlyeq 0, \mathbf {Tr}(\mathbf X) =
    1\}$ be the domain in which a feasible $\mathbf{X}$ lies. 
    Note that $\mathcal P$ is a convex set of $\mathbf{X}$.
    As shown in Line 4 in Algorithm~\ref{ALG:0}, we need to solve the
    following maximization problem:
    \begin{equation}
    \underset{\mathbf{X} \in \mathcal P} \max 
            \,\,\, {f(\mathbf X, \rho_k)},
    \end{equation}
    where $\rho_k$ is the output of Line 3 in Algorithm~\ref{ALG:0}.
    Our algorithm offers a simple and efficient way for solving
    this problem by explicitly maintaining the positive semidefiniteness property
    of the matrix $\mathbf{X}$. It needs only compute the largest eigenvalue
    and the corresponding eigenvector whereas most previous approaches such as the method of 
    \cite{Kilian}
    require  a full eigen-decomposition of $\mathbf{X}$.
    Their computational complexities are $ O( D^2) $ and $ O( D^3) $, respectively.
    When $ D $ is large, this computational complexity difference could be significant.

    Let $\nabla f(\mathbf{X}, \rho_k)$ be the gradient matrix of $f(\cdot) $ with
    respect to $\mathbf{X}$ and $\alpha$ be the step size for updating $\mathbf{X}$. 
    Recall that we update $\mathbf X$
    in such a way that $\mathbf X_{i+1}=(1-\alpha)\mathbf X_{i+1}+\alpha\triangle\mathbf X$, 
    where $\mathbf{rank}(\triangle\mathbf X)=1$ and $\mathbf{Tr}(\triangle\mathbf X)=1$. To find the
    $\triangle\mathbf X$ that satisfies these constraints and in the meantime can best approximate
    the gradient matrix $\nabla f(\mathbf{X}, \rho_k)$, we need to solve the following optimization
    problem:
    \begin{align}
        \label{EQ:Sub}
        \max_{\triangle\bf X} \quad & \big< \nabla f(\mathbf{X}, \rho_k),\triangle\mathbf X\big>
        \notag \\
     \st \,\,\,  & \mathbf{rank}(\triangle\mathbf X)=1, \mathbf{Tr}(\triangle\mathbf X)=1.
    \end{align}
    The optimal $\triangle{\mathbf X}^\star$
    is exactly $\mathbf v \mathbf v^{\T}$ where $\mathbf v$
    is the eigenvector of $\nabla f(\mathbf{X}, \rho_k)$ that
    corresponds to the largest eigenvalue.
    The constraints says that $ \triangle{\mathbf X} $ is a outer product
    of a unit vector:  $ \triangle{\mathbf X} = {\bf v} {\bf v}  ^\T $ with 
    $ || {\bf v} ||_2 = 1$. Here $ || \cdot ||_2 $ is the Euclidean norm. 
    Problem \eqref{EQ:Sub} can then be written as:
    $   \max_{\bf v}  {\bf v}^\T  [  \nabla f(\mathbf{X}, \rho_k) ] \bf v  $,
    subject to $ || {\bf v} ||_2 = 1$. It is clear now that an eigen-decomposition
    gives the solution to the above problem.

    Hence, to solve the above
    optimization, we only need to compute the leading eigenvector of
    the matrix $\nabla f(\mathbf{X}, \rho_k)$.
    Note that $\mathbf{X}$ still retains the
    properties of $\mathbf X \succcurlyeq 0, \mathbf{Tr}(\mathbf X) =  1$
    after applying this process.

    Clearly, a key parameter of this optimization process is $\alpha$ which
    implicitly decides the total number of iterations. The computational
    overhead of our algorithm is proportional to the number of iterations.
    Hence, to achieve a fast optimization process, we need to ensure
    that in each iteration the $\alpha$ can lead to a sufficient reduction on the
    value of $f$. This is discussed in the following part.

\subsection{Finding the optimal step size $\alpha$}\label{Sec:kNNMM-SDP2}

    We employ the backtracking line search algorithm in~\cite{backtracking}
    to identify a suitable $\alpha$. It reduces the value of $\alpha$ until
    the Wolfe conditions are satisfied. As shown in Algorithm~\ref{ALG:2},
    the search direction is $\mathbf{p}_i = \mathbf v_i \mathbf v_i^{\T} - \mathbf
    X_i$. The Wolfe conditions that we use are
        \begin{equation*}
        f(\mathbf{X}_i+\alpha\mathbf{p}_i, \rho_i)\leq f(\mathbf{X}_i,
        \rho_i)+c_1\alpha\mathbf{p}_i^{\T}\nabla
        f(\mathbf{X}_i, \rho_i),
        \end{equation*}
        \begin{equation}\label{eqn:armijo}
        \big|\mathbf{p}_i^{\T}\nabla f(\mathbf{X}_i+\alpha\mathbf{p}_i, 
        \rho_i)\big|\leq c_2\big|\mathbf{p}_i^{\T}\nabla f(\mathbf{X}_i,
        \rho_i)\big|,
        \end{equation}where $0 < c_1 < c_2 < 1$. The result of backtracking line search
    is an acceptable $\alpha$ which can give rise to sufficient reduction on
    the function value of $f(\cdot)$. We  show in the experiments that
    with this setting our optimization algorithm can achieve higher
    computational efficiency than some of the existing solvers.

\section{Experiments}
\label{Sec:experiments}

    The goal of these experiments is to verify the efficiency of our algorithm in achieving
    comparable (or sometimes even better) classification performances with a reduced computational cost.
    We perform experiments on $10$ data sets described in
    Table~\ref{Table:dataset}. 
    For some data sets, PCA is performed to remove noises and reduce the dimensionality. 
    The metric learning algorithms are then run on the data sets pre-processed by PCA.
    The Wine, Balance, Vehicle, Breast-Cancer and Diabetes data sets
    are obtained from UCI Machine Learning Repository \cite{UCI},
    and USPS, MNIST and Letter are from LibSVM \cite{LIBSVMdata2001}
    For MNIST, we only use
    its test data in our experiment. The ORLface data
    is from att research\footnote{http://www.uk.research.att.com/facedatabase.html} and
    Twin-Peaks is downloaded from L. van der Maaten's website\footnote{http://ticc.uvt.nl/lvdrmaaten/}.
    The Face and Background
    classes (435 and 520 images respectively) in the image retrieval experiment are obtained from the Caltech-101 object
    database \cite{Caltech-101}.
    In order to perform statistics analysis,
    the ORLface, Twin-Peaks, Wine, Balance, Vehicle, Diabetes and
    Face-Background data sets are randomly
    split as 10 pairs of train/validation/test subsets
    and experiments on those data set are repeated 10 times on each split.

\begin{table*}[t!]
\begin{center}
\caption{The ten benchmark data sets used in the experiment. Missing entries in ``dimension after
PCA'' indicate no PCA processing.}
\label{Table:dataset}
\centering
\begin{tabular}{ l|c|c|c|c|c|c|c|c }
\hline
& \# training &  \# validation &  \# test & dimension & dimension after PCA & \# classes  &
        \# runs   & \# triplets for SDPMetric\\
\hline\hline
USPS$_{\rm PCA}$    &   5,833    &   1,458    &   2,007    &   256 & 60  &   10  &   1   &  52,497   \\
        USPS  &5,833 &  1,458    &   2,007     &   256 &   &   10  &   1     &       5,833           \\
MNIST$_{\rm PCA}$    &  7,000    &   1,500    &   1,500    &   784 & 60  &   10  &   1   &  54,000         \\
MNIST   &  7,000    &   1,500    &   1,500    &   784 &  &   10  &   1      &    7,000 \\
        Letter  &   10,500  &   4,500   &    5,000   &    16 &    &   26  &  1       & 94,500     \\
        ORLface &   280 & 60 & 60 & ~2,576~& 42 & 40  &   10                         & 280    \\
        Twin-Peaks & ~14,000~ & ~3,000~ & ~3,000~ & 3& & 11 & 10                     & 14,000    \\
        Wine    &   126 & 26 & 26 & 13 & & 3   &   10                                & 1,134    \\
        Balance &   439 &   93  &   93  &   4  & &   3   &   10                      & 3,951    \\
        Vehicle &   593 &   127 &   126 &   18 & &   4   &   10                      & 5,337    \\
        Breast-Cancer  &   479 &   102 &   102 &   10 &  &   2   &   10              & 4,311    \\
        Diabetes  & 538 & 115 & 115 & 8 & & 2 & 10                                   & 4,842    \\
        Face-Background & 472 & 101 & 382 & 100 &  & 2 & 10                          & 4,428    \\
 \hline
\end{tabular}
\end{center}
\end{table*}

    The $k$-NN classifier with the Mahalanobis 
    distance learned by our algorithm (termed SDPMetric in short) is compared
    with the $k$-NN classifiers using a simple Euclidean distance
    (``Euclidean'' in short) and that learned by the Large Margin Nearest
    Neighbor in \cite{Kilian} (LMNN\footnote{In our experiment, we have used the implementation
    of LMNN's authors. Note that to be consistent with the
    setting in \cite{Kilian}, 
    LMNN here also uses the ``obj=1'' option and updates the projection matrix to
    speed up its computation. If we update the distance matrix directly to get global optimum,
    LMNN would be much more slower due to full eigen-decomposition at each iteration.}
    in short).
    Since  Weinberger \etal \cite{Kilian} has shown that LMNN obtains
    the classification performance comparable to support vector machines on some data sets,
    we focus on the
    comparison between our algorithm and LMNN, which is considered as the state-of-the-art.
    To prepare the training triplet set
    $\mathcal S$, we apply
    the $3$-NN method to these data sets and generate the training
    triplets for our algorithm. The training data sets for LMNN is also generated using $3$-NN,
    except that the Twin-peaks and ORLface are applied with the $1$-NN
    method. Also, the experiment compares the two variants of our proposed SDPMetric,
    which use the squared hinge loss (denoted as SDPMetric-S) and the Huber loss(SDPMetric-H), respectively.
    We split each data
    set into 70/15/15\% randomly and refer to those split sets
    as training, cross validation and test sets except pre-separated data
    sets (Letter and USPS) and Face-Background which was made for image retrieval.
    Following \cite{Kilian}, LMNN uses 85/15\%
    data for training and testing. The training data is also split
    into 70/15\% in LMNN for cross validation to be consistent with our SDPMetric.
    Since USPS data set has been split into training/test already, only the training data are
    divided into 70/15\% for training and validation. The Letter data set is separated
    according to  Hsu 
    and Lin \cite{CWH01a}. Same as in \cite{Kilian}, PCA is 
    applied to USPS, MNIST and ORLface to reduce the dimensionality
    of feature vectors.

    The following experimental study demonstrates that our algorithm achieves
    slightly better classification accuracy rates with a 
    much less computational cost
    than LMNN on most of the tested data sets. 
    The detailed test error rates and timing
    results are reported in Tables \ref{Table:error_rates}
    and~\ref{Table:time}.
    As we can see, 
    the test error rates of SDPMetric-S are comparable to those of LMNN. 
    SDPMetric-H achieves lower misclassification error rates than LMNN and the Euclidean distance
    on most of data sets except Face-Background data (which is treated as an image
    retrieval problem)
    and MNIST, on which SDPMetric-S achieves a lower error rate.
    Overall, we can conclude that the proposed SDPMetric either with squared hinge loss or
    Huber loss is at least comparable to (or sometimes slightly better than) 
    the state-of-the-art LMNN method in terms of classification performance.

\begin{figure}[b!]
        \centering
        \includegraphics[width=0.45\textwidth]{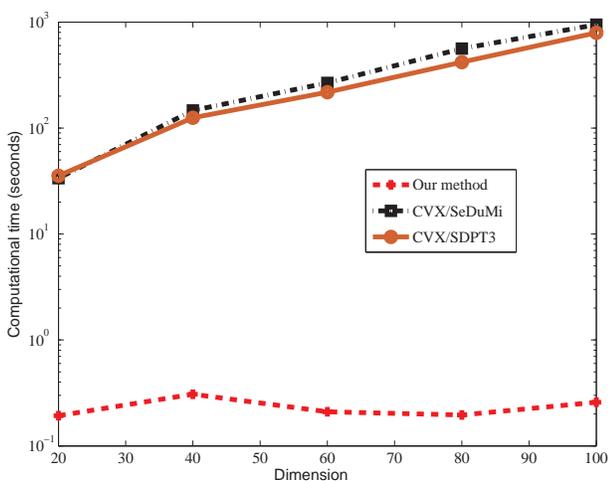}\\
        \caption{Computational time versus the dimensionality of feature vectors.
        } \label{fig:time}
        \end{figure}

    Before reporting the timing result on these benchmark data sets, we compared
    our algorithm (SDPMetric-H) with two convex optimization solvers, namely,
    SeDuMi \cite{Sturm2001SeDumi}
    and SDPT3 \cite{SDPT3}
    %
    which are used as internal solvers in the disciplined convex programming software CVX
    \cite{Grant2007CVX}.
    Both SeDuMi and SDPT3 use interior-point based methods. 
    To perform eigen-decomposition, 
    our SDPMetric uses ARPACK \cite{arpack},
    which is designed to solve large scale eigenvalue problems.
    Our SDPMetric is implemented in standard C/C++.
    Experiments have been conducted on a standard desktop. 
    We randomly generated $1,000$ training triplets and gradually increase the
    dimensionality of feature vectors from $20$ to $100$.
    Fig.~\ref{fig:time}
    illustrates computational time of ours, CVX/SeDuMi and CVX/SDPT3. As shown, the
    computational load of our algorithm almost keeps constant as the
    dimensionality increases. This might be because the proportion of eigen-decomposition's
    CPU time does not dominate with dimensions varying from $ 20 $ to $100$ in SDPMetric
    on this data set. 
    In contrast, the computational loads of CVX/SeDuMi
    and CVX/SDPT3 increase quickly in this course. In the case of the dimension of $100$,
    the difference on CPU time can be as large as $800 \sim 1000$ seconds.
    This shows the inefficiency and poor scalability of standard interior-point methods. 
    Secondly, the computational time of LMNN, SDPMetric-S and SDPMetric-H 
    on these benchmark data sets
    are compared in Table~\ref{Table:time}.
    As shown, LMNN is always slower than the proposed SDPMetric which converges very fast on
    these data sets. Especially, on the
    Letter and Twin-Peaks data sets, SDPMetric shows significantly
    improved computational efficiency.

\begin{figure}[ht]
        \centering
        \includegraphics[width=0.45\textwidth]
        {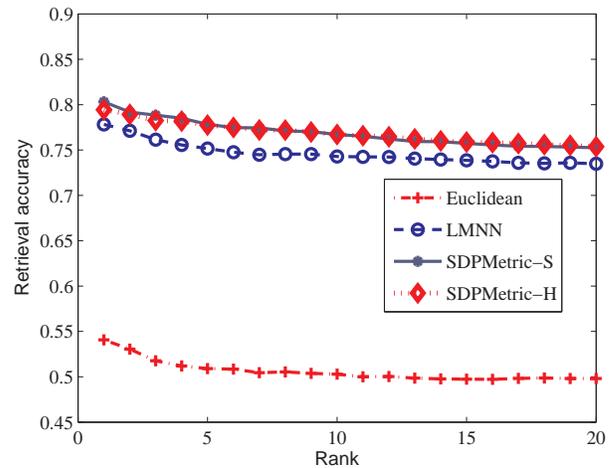}
        \caption{Retrieval performances of SDPMetric-S, SDPMetric-H,
        LMNN and the
        Euclidean distance. The curves of SDPMetric-S and SDPMetric-H are very close.}
        \label{fig:retrieval}
        \end{figure}

Face-Background data set consists of the two object classes,
Face-easy and Background-Google in \cite{Caltech-101}, as a
retrieval problem. The images in the class of Background-Google are
randomly collected from the Internet and they are used to represent
the non-target class. For each image, a number of interest regions
are identified by the Harris-Affine detector \cite{MikolajczykS04}
and the visual content in each region is characterized by the SIFT
descriptor \cite{lowe99object}. A codebook of size $100$ is created by
using $k$-means clustering. Each image is then represented by a
$100$-dimensional histogram vector containing the number of occurrences of
each visual word. We evaluate retrieval accuracy using each facial
image in a test subset as a query. For each compared metric, the
{\bf accuracy} of the retrieved top $1$ to $20$ images are
computed, which is defined as the ratio of the number of facial
images to the total number of retrieved images. We calculate the
average accuracy of each test subset and then average over the whole
$10$ test subsets. Fig.~\ref{fig:retrieval} shows the retrieval
accuracies of the Mahalanobis distances learned by Euclidean, LMNN
and SDPMetric. Clearly we can observe that SDPMetric-H and SDPMetric-S
consistently present higher retrieval accuracy values, which again
verifies their advantages over the LMNN method 
and Euclidean distance.

\begin{table*}[ht]
\begin{center}
\caption{$ 3$-Nearest Neighbor misclassification error rates. The standard deviation
values are in brackets. The best results are highlighted in bold.}
\label{Table:error_rates} \centering
 \begin{tabular}{l|c|c|c|c}
        \hline
                & Euclidean & LMNN  & SDPMetric-S   & SDPMetric-H  \\
                \hline \hline
USPS$_{\rm PCA}$    & 5.63  & \textbf{5.18}      &   5.28    &  \textbf{5.18}\\
USPS            & 4.42  & 3.58      &   4.36    &  \textbf{3.21}\\
MNIST$_{\rm PCA}$    & 3.15  &       3.15    &       \textbf{3.00}    &       3.35 \\
MNIST          & 4.80  &       4.60    &       \textbf{4.13}    &       4.53 \\

        Letter  & 5.38 & 4.04 &  3.60 &  \textbf{3.46}\\
        ORLface & 6.00 (3.46) & 5.00 (2.36) &  4.75 (2.36)  & \textbf{4.25 (2.97)}\\
        Twin-Peaks & 1.03 (0.21) & 0.90 (0.19) & 1.17 (0.20) & \textbf{0.79 (0.19)}\\
        Wine    & 24.62 (5.83) & 3.85 (2.72)  &    3.46 (2.69)  &  \textbf{3.08 (2.31)}\\
        Bal     & 19.14 (1.59) & 14.19 (4.12) &  \textbf{9.78 (3.17)}   &  10.32 (3.44) \\
        Vehicle & 28.41 (2.41) & 21.59 (2.71) &  21.67 (4.00) &   \textbf{20.87 (2.97)} \\
        Breast-Cancer  & 4.51 (1.49) & 4.71 (1.61)  &  3.33 (1.40)  &  \textbf{2.94 (0.88)} \\
        Diabetes  & 28.00 (2.84) & 27.65 (3.45) & 28.70 (3.67) & \textbf{27.64 (3.71)}\\
        Face-Background & 26.41 (2.72) & \textbf{14.71 (1.33)} & 16.75 (1.72) & 15.86 (1.37) \\
        \hline

\end{tabular}
\end{center}
\end{table*}

\begin{table}[h!]
\begin{center}
\caption{Computational time for each
run.}\label{Table:time}
        \begin{tabular}{l|c|c|c}
        \hline
            &   LMNN    &   SDPMetric-S   &   SDPMetric-H \\ \hline \hline
USPS$_{\rm PCA}$    &   256s  &  111s  &  258s \\
USPS                &   1.6h  &  16m  &  20m \\
MNIST$_{\rm PCA}$   &   219s  &  111s  &  99s  \\
MNIST               &   9.7h  &  1.4h  &  37m  \\
        Letter  &   1036s  &    6s  &   136s \\
        ORLface &   13s &   4s  &   3s \\
        Twin-peakes & 595s & less than 1s & less than 1s \\
        Wine    &   9s  &   2s  &   2s \\
        Bal &   7s  &   less than 1s  & 2s \\
        Vehicle &   19s &   2s  &   7s \\
        Breast-Cancer  &   4s  &   2s  &   3s \\
        Diabetes  &   10s  &   less than 1s  &   2s \\
        Face-Background & 92s & 5s & 5s \\
        \hline
\end{tabular}
\end{center}
\end{table}

%
%

\section{Conclusion}\label{Sec:conclusion}

    We have proposed a new algorithm to demonstrate how to efficiently learn a
    Mahalanobis distance metric
    with the principle of margin maximization.
    Enlightened by the important theorem on \psd matrix decomposition in
    \cite{Shen2008psd}, we
    have designed a gradient descent method to update the Mahalanobis matrix with cheap computational
    loads and at the same time, the \psd property of the learned matrix is maintained
    during the whole optimization process. Experiments on
    benchmark data sets and the retrieval problem verify the superior classification performance and
    computational efficiency of the proposed distance metric learning algorithm.

    The proposed algorithm may be used to solve more general SDP problems in machine learning.
    To look for other applications is one of the future research directions. 


\end{document}